# Model Based Framework for Estimating Mutation Rate of Hepatitis C Virus in Egypt

Nabila Shikoun
Ph.D. Student
Faculty of Engineering
Al Azhar University
Email: nabila_shikoun@yahoo.com

Mohamed El Nahas
Prof. of Pattern Recognition
Faculty of Engineering
Al Azhar University
Email: mel_nahas@hotmail.com

Samar Kassim
Prof. of Medical Biochemistry & molecular Biology
Faculty of Medicine
Ain Shams University
Email: samar_kassim@yahoo.com

*Abstract* - Hepatitis C virus (HCV) is a widely spread disease all over the world. HCV has very high mutation rate that makes it resistant to antibodies. Modeling HCV to identify the virus mutation process is essential to its detection and predicting its evolution. This paper presents a model based framework for estimating mutation rate of HCV in two steps. Firstly profile hidden Markov model (PHMM) architecture was builder to select the sequences which represents sequence per year. Secondly mutation rate was calculated by using pair-wise distance method between sequences. A pilot study is conducted on NS5B zone of HCV dataset of genotype 4 subtype a (HCV4a) in Egypt.

Keywords: Hepatitis C virus (HCV), Profile Hidden Markov Model (PHMM), Non-structure 5 B(NS5B), Phylogenetic tree, pair-wise distance.

## I. INTRODUCTION

Once hepatitis C virus (HCV) has been discovered, it has been an importance subject of research and clinical investigations as its major role in human disease has emerged. An estimated 170 million people (3% of the world's population) worldwide have hepatitis C virus (HCV) infection and creates a huge disease burden from chronic, progressive liver disease [1]. Hepatitis C is a predominant genotype found throughout the Middle East and parts of Africa, with high population prevalence in Egypt. Due to the world's constant effort to find treatment for this fatal disease; many researches and trials have been made [2]. HCV has become a major cause of liver cancer and one of the commonest indications for liver transplantation. HCV infection can be treated, but this is costly and requires long-term medical support and follow-up; current therapies are impractical for the majority of HCV carriers worldwide. The development of a protective vaccine remains, a distant goal [1].

The aim of this study is to identify evolution model which estimates the mutation rate of non-structure 5B (NS5B) which has more variation (mutation) than other zones in Hepatitis C Virus genotype 4 subtype a (HCV4a) in Egypt. Profile hidden markov model (PHMM), phylogenetic tree, a polynomial fitting and a least-square interpolation is used to outline the progression of the viral mutations over time. This model shall be used to predicate mutation rate of HCV in blood samples. The mutation model presents new therapeutic targets as well as genomic information for designing vaccine candidates.

In this research the profile hidden markov model (PHMM) was identified as a technique to select sequence from many sequences observed in one year. A relationship between several sequences was deduced representing several years by estimating mutation rate using pair-wise distance method. The mutation model can predicate mutation rate for next year. Our approach introduces NS5B zone in HCV4a genome for 5 years from 2007 to 2010.

This paper is organized as follows: In section 2, related research for evolution of hepatitis C virus is presented. Section 3 describes HCV virus genome and explains the concept of the mutation rate. Section 4 presents the suggested method for evaluation of mutation rate. Section 5 presents the study data set and experimental results. Section 6 concludes the paper with future research directions.

## II. RELATED RESEARCH

Several researches have been conducted to unravel information and useful patterns in a database for evolution of RNA and calculating mutation rate in HCV virus. Pybus et al. [3] developed Bayesian inference framework to estimate the transmission dynamics of HCV in Egypt from sampled viral gene sequences, and to predict the public health impact of the virus. Bruijne et al. [4] investigates the genetic diversity and evolutionary origin of HCV-4 in Amsterdam, The Netherlands used a molecular epidemiological approach. Phylogenetic analysis of the NS5B sequences (668 bp) was obtained from 133 patients newly diagnosed with HCV-4 infection over the period from 1999 to 2008. Xiong et al. [5] propose a stochastic model based on the branching process for estimation and comparison of the mutation rates in proliferation processes of cells or microbes. Barash et al. [6] introduce some Programs for RNA mutational analysis. These programs can be used for suggesting point mutations, investigating the effect of deleterious and compensatory mutations in allosteric ribozymes and riboswitches and analyzing regulatory RNA sequences by their mutational profile. Ribeiro et al. [7] measured the accumulative rate of mutations and fitted the model to the sequence data of HCV by estimating the median in vivo viral mutation rate.

## III. HCV VIRUS GENOME

The HCV genome is an enveloped structure approximately 50 nm in diameter. HCV is a positive single stranded enveloped RNA virus belonging to the *Flaviviridae* family with an average length of 9600 bases and carries a single, long open reading frame (ORF) flanked by 5' and 3' non-translated regions. The ORF encodes a polyprotein of ~ 3000 amino acids that is processed into three structural proteins (Envelopes 1 and 2 and p7) and six non-structural proteins named NS2-NS5B [8 -9].

HCV is classified into eleven major genotypes (designated 1-11), many subtypes (designated a, b, c, etc.), and about 100 different strains (numbered 1, 2, 3, etc.) based on the genomic sequence heterogeneity. Genotype 4 (HCV4) is principally found in the Middle East and Africa, particularly Egypt, which represent more than 90% of infections due to genotype 4 worldwide [9]. Ray et al. [10] characterize the genotype distribution of HCV in Egypt. In their work, specimens were obtained from blood donors in 15 geographically diverse governorates throughout Egypt. The result showed that 111 (91%) were genotype 4, 1 (1%) was genotype 1a, 1 (1%) was genotype 1b, and 9 (7%) could not be typed.

**Definition of Mutation Rate**

A mutation is a change of the nucleotide sequence of the genome of an organism and virus. Mutation rate is a measure of the rate at which various types of mutations occur during some unit of time [11]. There are two units used in viral mutation rate. These units are related with the nature of the viruses replication. Viruses replication are one of two types: binary replication and Linear replication "stamping machine". In case of binary replication, mutation rates are expressed as substitutions per nucleotide per strand copying unit. For linear replication mutation rates are expressed as substitutions per nucleotide per cell infection unit. HCV mutation rate is expressed with the last units [12]. In this paper mutation rate as substitutions per nucleotide per cell infection unit time was adopted. Reasons of mutations are from unrepaired damage to DNA or to RNA genomes (caused by radiation or chemical mutagens), from errors in the process of replication, or from the insertion or deletion of segments of DNA by mobile genetic elements. The high mutation rate in HCV is the reason of persistence in the human host [11].

An accurate estimate of mutation rate of virus is very important to understand the evolution of the viruses and to combat them [12]. There are various statistical methods to estimate mutation rates. It's classified into three general approaches: linear regression, maximum likelihood, and Bayesian inference. Linear regression procedures calculate substitution rate by comparing directly genetic distance between two sequences with the interval separating their isolation times. These methods are fast and useful for visualizing new data sets. They can assist in model selection. But they make several limiting assumptions. Maximum likelihood (ML) approach defines methods that accommodate the time structure of temporally spaced sequences. ML methods are more sensitive and accurate than distance-based methods of linear regression. In Bayesian statistical inference, substitution rate is obtained empirically from the frequency distribution of the parameters values sampled by the Markov chain Monte Carlo (MCMC) algorithm. Maximum likelihood and Bayesian methods utilize more information from the sequences and allow much more complex models of molecular evolution and demography to be investigated [13].

One of linear regression methods is based on pair-wise distance [13]. Pair-wise distance enables to estimate distances in terms of the number of nucleotide substitutions. In this paper two types of distances to measure the genetic distance between sequences are used: Jukes Cantor and Kimura.

**Jukes & Cantor distance**: assumes all changes between all nucleotide are equally [14]. This is defined in equation (1)

$$d_{xy} = -(3/4)\log_e(1 - 4/3\,D) \qquad (1)$$

$d_{xy}$ = distance between sequence x and sequence y expressed as the number of changes per site,

D = is the observed proportion of nucleotides which differ between two sequences (fractional dissimilarity).

The 3/4 and 4/3 terms reflect that there are four types of nucleotides and three ways in which a second nucleotide

**Kimura method**: this method distinguishes between two types of differences when comparing a pair of nucleotide sequences. Transition type which gets distance difference between nucleotide both are purines or both pyrimidines (T↔C, A↔G). The second type is transversion where the difference distance between one of the two is a purine and the other is a pyrimidine (T↔A, T↔G, C↔A, C↔G) [15]. This method is defined in equation (2)

$$K = -\tfrac{1}{2}\log_e\left\{(1-2P-Q)\sqrt{1-2Q}\right\} \qquad (2)$$

Where P and Q are the fractions of nucleotide sites of transition and transversion types respectively between two compared sequences.

## IV. EVOLUTION OF MUTATION RATE

The objective of evolution model is to estimate mutation rate using Profile Hidden Markov Model (PHMM). Where PHMM is a special type of left-to-right HMMs is commonly used to model multiple alignments. The architecture of PHMM was introduced by Krogh (1994) [16]. El Nahas et al. [9] define a tuple of HMM, its tasks, and the architecture of PHMM.

The first step is to collect sequences from Egypt for several years. The data collected from region NS5B sequences from HCV genotype 4 subtype a (HCV4a). Choice one sequence represent certain year from sequences collected to this year. This is done by using PHMM. The Baum Welch algorithm is generally accepted to estimate PHMM parameters. However, this algorithm assumes that the model length is known, which not the case in this work. Hence, we have to adapt the learning procedure to search for the optimal model length. The second step gathering all choice sequences and get matrix distance between each two sequences. Drawing these genetic distances and fitting it to get polynomial mutation rate. In this work MATLAB bioinformatics tool box functions was uses. The procedure is detailed in the following steps:

**Procedure Evaluate Mutation Rate**
**Begin**
1. **Input** all sequences of each year of NS5B region of HCV4a.
2. **Apply Multiple Sequence Alignment (MSA)** to these sequences of each year. MSA performs by using a heuristic search known as progressive technique (also known as the hierarchical or tree method).
3. **Preprocess data to filter unknown symbols**. Sometimes, a characters 'r, g, w, m, n, y, k, s' is found in sequences which do not map to any of 4 nucleotides (A, C, G, T) so it is replaced by one of nucleotide by using MSA.
4. **Initialize structure for PHMM of MSA**. The initial model structure and length are defined using information derived from the alignment together with its prior knowledge of the general nature of proteins.
5. **Estimate the PHMM parameters from training sequences to each year**. All the parameters in the PHMM (i.e. the transition probabilities and the nucleotide distributions) are estimated from a set of aligned sequences to maximize the likelihood of the observed sequences in the family. The likelihood of observed sequences is defined as:

    P(sequences | model) = P(sequence 1| model) * … * P(sequence n |model)

6. **Score the model**. Scoring is used to assign a score with respect to the model to any query sequence for each year, the better the score, the higher and the chance that the query sequence is a member (homologue) of the protein family represented by the model. Scores are computed using log-odd ratios for emission probabilities and log probabilities for state transitions.
7. **Computing the sequence Pair-Wise Distances**. That is by constructing distance matrix which computes distances between each sequence pair. The method to calculate pair-wise distances is 'Jukes-Cantor', which estimate Maximum likelihood of the number of substitutions between two sequences. Ignore sequence sites representing gaps. P is described with the method P-distance. Proportion of sites at which the two sequences are different. Note that P is close to 1 for poorly related sequences, and P is close to 0 for similar sequences.
8. **Construct a phylogenetic tree**. Using distance matrix computed in step 7 to build the phylogenetic tree. Where Phylogenetic trees represent evolutionary relationships, or genealogy, among species. The neighbor-joining method used to build the tree. Assuming equal variance and independence of evolutionary distance estimates.
9. **Estimate the date of origin of the mutation**. Consider the pair-wise distances according to the Kimura method, which distinguishes between transitional and tranversional mutation. Then, restrict analysis to the distance of each sequence from the reference year consider as the starting of the mutation. Plot the genetic distance versus the date of collection.
10. **Compute Progression of Viral Mutation**. Perform a polynomial fitting and a least-square interpolation to outline the progression of the viral mutations over time. Which finds the coefficients of a polynomial `p(x)` of degree `n` that fits the data, `p(x(i))` to `y(i)`, in a least squares sense. The result `p` is a row vector of length `n+1` containing the polynomial coefficients in descending powers.
11. Reroute the Phylogenetic Tree. The rerouted tree better illustrates the progression of the NS5B(HCV4a) mutation starting with the early infections.

**end**

## V. PILOT STUDY

The main objective of this pilot study is to estimate the mutation rate of region NS5B in HCV4a spread in Egypt. Genotype 4 (HCV4) is particularly principally found in Egypt, which represent more than 90% of infections worldwide [12]. The region of NS5B in HCV4 subtype a (HCV4a) is used for this purpose. A data set representing NS5B is collected and used for to identify its mutation rate. Then, the learning procedure described in preceding section is applied on this real world data set to identify the model.

### A. Data Description

The dataset contains the genomic sequences of distinct non-structural proteins named NS5B genotype 4 subtypes 4a from Egypt (11 sequences) in 2007 year, (35 sequences) in 2008 year, (35 sequences) in 2009 year and (17 sequences) in 2010 year. The length of the sequences varied between 286 and 339 nucleotide (nt). The data is obtained from the site "http://www.ncbi.nlm.nih.gov/protein" and it is presented in Table 1 which contains virus sequence Number (Virus_seq_no) and genebank of each year.

The data set was found to contain the characters 'r, g, w, m,n, y, k, s' which are undefined as nucleotide, to overcome this problem and to replace these undefined sites with suitable nucleotide the following steps were applied

1. Determine sequence number and positions numbers which contain these characters in original sequences.

2. Apply global alignment to all sequences MSA, and determine sequence number, position number of these characters and the most character repeated

3. Replace these character with suitable nucleotide found in step 2.

**TABLE 1: The Data Set** (http://www.ncbi.nlm.nih.gov/protein)

| Virus_seq_no | Genbank_2007 | Virus_seq_no | Genbank_2008 | Virus_eq_no | Genbank_2009 | Virus_seq_no | Genbank_2010 |
|---|---|---|---|---|---|---|---|
| 2007_1 | DQ911222 | 2008_1 | EF694448 | 2009_1 | AB470254 | 2010_1 | FN668600 |
| 2007_2 | DQ911173 | 2008_2 | EF694425 | 2009_2 | AB470018 | 2010_2 | FN668570 |
| 2007_3 | DQ911190 | 2008_3 | EF694438 | 2009_3 | AB470024 | 2010_3 | FN668577 |
| 2007_4 | DQ911228 | 2008_4 | EF694446 | 2009_4 | AB470055 | 2010_4 | FN668587 |
| 2007_5 | DQ911183 | 2008_5 | EF694408 | 2009_5 | AB470015 | 2010_5 | FN668591 |
| 2007_6 | DQ911164 | 2008_6 | EF694420 | 2009_6 | AB470046 | 2010_6 | FN668593 |
| 2007_7 | DQ911172 | 2008_7 | EF694424 | 2009_7 | AB470009 | 2010_7 | FN668574 |
| 2007_8 | DQ911208 | 2008_8 | EF694455 | 2009_8 | AB470048 | 2010_8 | FN668588 |
| 2007_9 | DQ911205 | 2008_9 | EF694502 | 2009_9 | AB470030 | 2010_9 | FN668589 |
| 2007_10 | AY548725 | 2008_10 | EF694517 | 2009_10 | AB470036 | 2010_10 | FN668586 |
| 2007_11 | DQ911206 | 2008_11 | EF694456 | 2009_11 | AB470019 | 2010_11 | FN668583 |
| 2007_12 | AY548723 | 2008_12 | EF694413 | 2009_12 | AB470060 | | |
| 2007_13 | DQ911231 | 2008_13 | EF694525 | 2009_13 | AB470057 | | |
| 2007_14 | DQ911163 | 2008_14 | EF694501 | 2009_14 | AB470021 | | |
| 2007_15 | DQ911207 | 2008_15 | EF694513 | 2009_15 | B470039 | | |
| 2007_16 | AY548722 | 2008_16 | EF694498 | 2009_16 | AB470028 | | |
| 2007_17 | DQ911218 | 2008_17 | EF694396 | 2009_17 | AB470008 | | |
| | | 2008_18 | EF694486 | 2009_18 | AB470014 | | |
| | | 2008_19 | EF694450 | 2009_19 | AB470037 | | |
| | | 2008_20 | EF694418 | 2009_20 | AB470027 | | |
| | | 2008_21 | EF694439 | 2009_21 | AB470052 | | |
| | | 2008_22 | EF694434 | 2009_22 | AB470043 | | |
| | | 2008_23 | EF694442 | 2009_23 | AB470049 | | |
| | | 2008_24 | EF694440 | 2009_24 | AB470026 | | |
| | | 2008_25 | EF694493 | 2009_25 | AB470006 | | |
| | | 2008_26 | EF694499 | 2009_26 | AB470053 | | |
| | | 2008_27 | EF694445 | 2009_27 | AB470013 | | |
| | | 2008_28 | EF694463 | 2009_28 | AB470023 | | |
| | | 2008_29 | EF694505 | 2009_29 | AB470005 | | |
| | | 2008_30 | EF694476 | 2009_30 | AB470007 | | |
| | | 2008_31 | EF694475 | 2009_31 | AB470056 | | |
| | | 2008_32 | EF694433 | 2009_32 | AB470017 | | |
| | | 2008_33 | EF694478 | 2009_33 | AB470059 | | |
| | | 2008_34 | EF694459 | 2009_34 | AB470252 | | |
| | | 2008_35 | EF694524 | 2009_35 | AB470033 | | |

## B. Experimental Results

Table 2 shows scores of PHMM model applied to all sequences of 2007 to 2010 years on NS5B of HCv4a. Scores show that the sequences 11, 9, 14 and 10 are more suitable candidate sequences of 2007, 2008, 2009 and 2010 years respectively.

**TABLE 2: Scores of sequences in each year to PHMM model**

| | | 2007 | 2008 | 2009 | 2010 |
|---|---|---|---|---|---|
| | | Seq_11 | Seq_9 | Seq_14 | Seq_10 |
| Number of Match states (length of the model) | 200 | 173.2023 | 290.6751 | 305.7761 | 112.8878 |
| | 210 | 189.8152 | 317.9683 | 318.7579 | 122.2278 |
| | 220 | 206.9295 | 334.3079 | 328.1445 | 126.6255 |
| | 230 | 223.6203 | 350.1565 | 342.7904 | 132.7961 |
| | 240 | 240.3303 | 363.4668 | 356.4055 | 153.8358 |
| | 250 | 254.2443 | 377.1032 | 369.6046 | 170.5305 |
| | 260 | 270.6948 | 388.6816 | 385.3829 | 189.7923 |
| | 270 | 288.0402 | 402.3539 | 394.7695 | 209.4087 |
| | 280 | 305.6335 | 415.4973 | 405.1091 | 228.9011 |
| | 290 | 321.3509 | 425.3251 | 418.0684 | 249.4041 |
| | 300 | 337.0683 | 436.9035 | | 269.9071 |
| | 310 | 353.6787 | 446.3764 | | 290.6723 |
| | 320 | 365.4927 | 457.7152 | | 301.4251 |
| | 330 | 369.5766 | | | |

Table 3 and Figure 1 show the scores of pair-wise distances using Juke-cantor correction expressed in a symmetric matrix.

**TABLE 3: Pair-Wise Distances using Jack-cantor corrections**

| | | 2007 | 2008 | 2009 | 2010 |
|---|---|---|---|---|---|
| **Years** | **2007** | 0 | 0.1077 | 0.1942 | 0.1009 |
| | **2008** | 0.1077 | 0 | 0.184 | 0.0579 |
| | **2009** | 0.1942 | 0.184 | 0 | 0.1956 |
| | **2010** | 0.1009 | 0.0579 | 0.1956 | 0 |

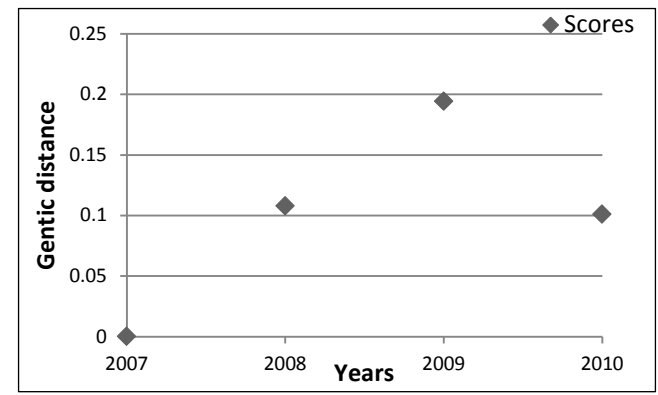

**Figure 1** Construct score of genetic distances versus years

Figure 2 shows a phylogenetic tree of Table 3. The distance difference between 2008 and 2010 years is 0.0579 which is the least distance record in Table 3. Hence these two years are represented as on cluster. Year 2009 records high difference distance with respect to all years, so it is considered as one cluster as shown in Figure 2.

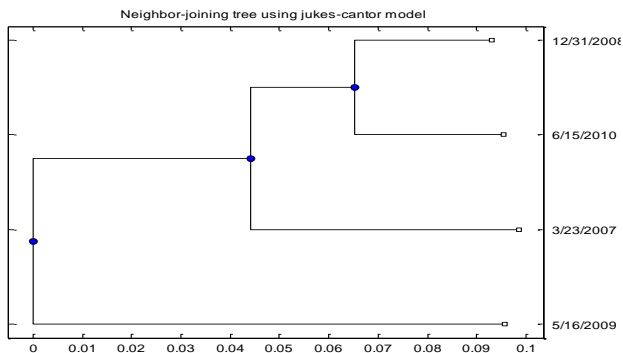

**Figure 2** Neighbor-Joining Phylogenetic tree

Table 4 shows pair-wise distances using Kimura method. Estimate the date of origin of the mutation is 3/23/ 2007. Restrict analysis to the distance of each sequence to this year. Figure 3 plots the genetic distance versus this year, which shows that the minimum distance recorded is 0.0424 to 2009 year respect to 2007 year. The maximum distance recorded is 0.0589 to 2008 year with respect to 2007 year.

**TABLE 4: Pair-Wise Distances using Kimura model**

| | | 2007 | 2008 | 2009 | 2010 |
|---|---|---|---|---|---|
| **Years** | **2007** | 0 | 0.0589 | 0.0424 | 0.0515 |
| | **2008** | 0.0589 | 0 | 0.0288 | 0.0584 |
| | **2009** | 0.0424 | 0.0288 | 0 | 0.0397 |
| | **2010** | 0.0515 | 0.0584 | 0.0397 | 0 |

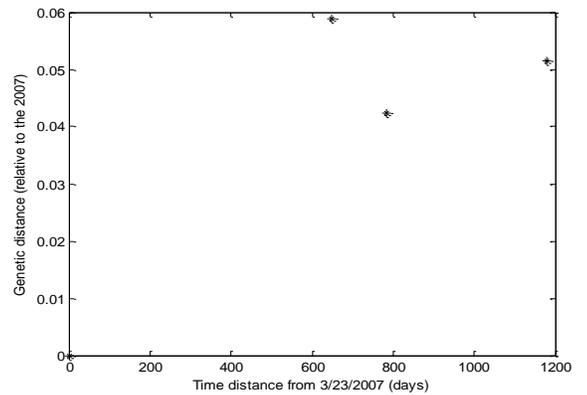

**Figure 3** Genetic distances with respect to the year of 2007

Figure 4 shows a polynomial fitting of progression of viral mutation from scores in genetic distance in Figure 3. These scores increase approximately in a linear manner with time. The polynomial equation is found to be:

$$D = 0.00004493\,X + 0.008826 \quad (5.1)$$

where 0.00004493 is the mutation rate(slope), 0.008826 is the intercept,

and D the is any genetic distance (score) referred to year of 2007 and X is the time relative to the reference year.

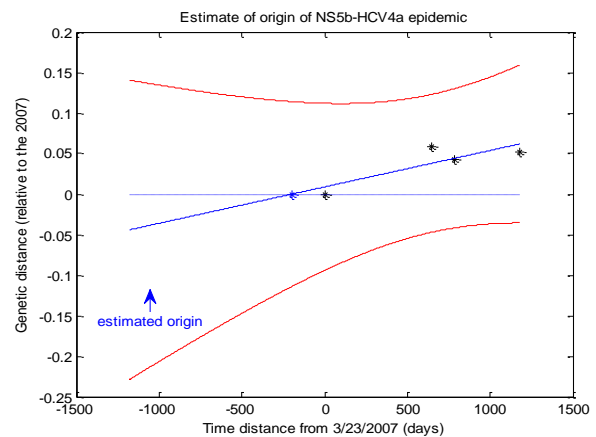

**Figure 4** Progression of Viral Mutation with respect to the year of 2007

Figure 5 shows rerouting the phylogenetic tree of NS5B(HCV4a). This illustrates the progression of the mutation of the virus. 95% confidence interval of the estimated polynomial is also plotted in red color on the Figure 5.

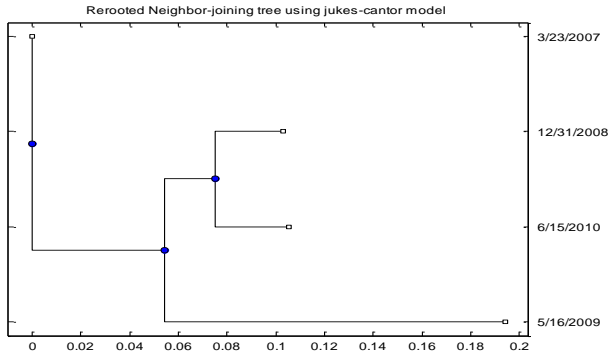

**Figure 5** Rerouting Neighbor-Joining Phylogenetic tree

## VI. CONCLUSION AND FUTURE WORK

In this paper, the model based framework for estimation of mutation rate of NS5B region of (HCV4a) in Egypt for several years is introduced. The framework is based on building phylogenetic tree from time tagged PHMM models. A learning process of PHMM model is used to select one sequence from all sequences collected in any year. Then scores based on genetic distance between any two sequences of selected sequences is calculated. Estimation of mutation rate of selected sequences is calculated by fitting the scores refereed to the 2007 year. The mutation rate is estimated as 0.00004493 substitutions per nucleotide per year. This approach combines both the speed of regression methods and the accuracy of statistical methods for evaluation of virus mutation rate. The future work shall study the mechanism of the virus mutation between all sequences and between the positions.